# Enhancing the Performance of Semantic Search in Bengali using Neural Net and other Classification Techniques

**Arijit Das[1], Diganta Saha[1]**

[1]Department of Computer Science and Engineering, Faculty of  Engineering and Technology,

Jadavpur University, Kolkata, West Bengal, 700032 INDIA

Corresponding author: Arijit Das (e-mail: arijit.das@ieee.org).

*Abstract Search has for a long time been an important tool for users to retrieve information. Syntactic search is matching documents or objects containing specific keywords like user-history, location, preference etc. to improve the results. However, it's often possible that the query and the best answer have no term or very less number of terms in common and syntactic search can't perform properly in such cases. Semantic search, on the other hand, resolves these issues but suffers from lack of annotation, absence of WordNet in case of low resource languages. In this work, we have demonstrated an end to end procedure to improve the performance of semantic search using semi-supervised and unsupervised learning algorithms. An available Bengali repository was chosen to have seven types of semantic properties primarily to develop the system. Performance has been tested using Support Vector Machine, Naive Bayes, Decision Tree and Artificial Neural Network (ANN). Our system has achieved the efficiency to predict the correct semantics using knowledge base over the time of learning. A repository containing around million sentences, a product of TDIL project of Govt. of India, was used to test our system at first instance. Then the testing has been done for other languages. Being a cognitive system it may be very useful for improving user satisfaction in e-Governance or m-Governance in the multilingual environment and also for other applications.*

## I.  INTRODUCTION

Semantic Search has been around for quite some time and has gained widespread use due to its applications and promising results. Most of the developing countries are multilingual. The emerging economies of the world also use more than one official language for communication.

India, the largest multilingual democracy, has 22 languages which have official recognition in the constitution and gets encouragement from the government to promote. In India, there are also 122 languages which are being spoken by more than ten thousand people and defined as major languages. Except these 1599 other languages also exist in India which are used by a very small portion of the population. India has seventy percent rural population and the majority of them are only proficient in their mother tongue. They prefer to use native language over the internet or in another way it can be said that they use the internet more for all e-Governance application if the content is available in their mother language. Search is one of the major operations which is done frequently by internet users. Let's see some case studies where "Semantic Search" or "Contextual Meaning" prevails in case of different language domain.

Let some Bengalee person (people of West Bengal, India or Bangla Desh whose mother tongue is Bengali) needs to reset his watch, so he wants to know the accurate time over the web and gives a search "কটা বাজে?"/kaṭā bāje?/ "What is the time now?" As of 06.07.2019 at 15:43 google, bing, yahoo all fail to give the answer either they are showing blank result or giving some pages which have the term "কটা বাজে? /kaṭā bāje?/". But the searcher who does not know the English language (let) wants to know the time, so search result should include local time, GMT etc. The search engine needs to understand the meaning or context of the searchers' search query. For a smart search engine such queries should point to the same answer for the query "what is the time now?" but search engines fail to understand the meaning of the query, therefore, cannot retrieve the current local time or Greenwich Mean Time.

Citizens' feedback is one of the most important pillars of good governance. E-governance makes the task of giving feedback easy and affordable. Giving input in the native language is easy nowadays with the soft keyboard available in their native languages. But if a question is asked in the Nepali language and the answer is present in the Portuguese language, the system fails to retrieve the result. As a case study, suppose a farmer of Darjeeling district of West Bengal is asking a question over the internet about the orange farming in Nepali and the answer is already present in Portuguese. Due to the lack of common words the search engine fails to populate the correct answers.

Meaning of the word changes with its use in the sentence in any language. Word Sense Disambiguation (WSD) is used to differentiate the actual meaning of the same word used differently in different texts. For example, the word "Bank" represents different meaning leading to different senses in various contexts or senses. The word "Bank" can point a financial organization, a riverside or seaside, a proper noun, a common noun even a verb or an adverb. It is difficult for a machine to differentiate the context which is easy for a human being with his or her innate linguistic intelligence. The branch of Word Sense Disambiguation (WSD) focuses on this challenge where system or machine is trained in such a way that it becomes able to differentiate the meaning of the same word used in different contexts. The method of learning for a machine may be statistical formulae or grammatical rules or consultation of dictionaries or WordNet for the meaning or sense of neighbor words.

The way of machine learning to predict the semantics can be supervised, unsupervised or semi-supervised. In case of supervised learning, the system predicts based on some predefined rules. Thus framing these rules is tedious and time-consuming. In the case of unsupervised learning, the system learns to predict from the past prediction and accuracy increases over time. It uses a series of statistical algorithms. The result of the system improves over time and not accurate in the first instances. But here human labor and time consumption are much less. Semi-supervised technique tries to take advantage of both the "supervised" and "unsupervised" method.

We have taken a repository of nearly one million sentences, which was taken from ISI Kolkata funded by Ministry of Electronics and IT, Govt. of India as a project named TDIL (Technology Development for Indian Languages). If the answers are available for the query, it was returned, no matter whether there is any common term between question (sentence-1) and the retrieved answer (sentence-2) or not. Two examples of sentence-1 or the questions are:

কে এবার আইপিএল এ সবচেয়ে বেশী রান করেছে?
(/ke ebar IPL e sabcheye beshee run korechhe?/ or "Who has scored highest in this IPL?")

রোনাল্ডো কোন ফুটবল ক্লাব এর সাথে যুক্ত? (/Ronāldo kon football club er sāthe jukta?/ or "With which football club Ronaldo is associated?")
Our system is returning the correct answer.
A set of 250 questions were fired and answers were collected and the final result has been evaluated by the experts.

## II. RELATED WORK

[1] contains a review of various measures for semantic similarities and review of various measures, such as- art measurement, feature-based measurement, measurement-based on length of path, information-based solution etc.
Iglesias et al. (2018) proposed method 'wpath' to combine the two conventional measuring techniques: information content and path length based measure to measure semantic similarity in Knowledge based Graphs (KGs) and DBPedia. The proposed method has an improvement than other measuring methods when performed over a well known dataset [2].

Semantic similarity plays a major role for the retrieval of information and web mining to retrieve semantically similar documents with the query submitted by the user [3]. The proposed method uses synset, a new method to calculate the similarity between terms, where online resources are used to derive the synsets. The benefit of the introduced work is that, semantic equivalence is computed among words, which helps to convert a query with query suggestions or most suitable queries.

Some meaningful similarity and related methods have already been developed. Various similarities and related methods have been proved useful in certain applications of computational intelligence. These methods are generally classified into four groups: path-length-based methods, depth based measure, feature-based methods and information-based methods. Path-length-based approach is another natural and direct way to evaluate meaningful similarity in ontology. Depending on the length of the path, the representative path involves the least path in the meaningful similarity measures. Wu and Palmer measure [4], Leacock and Chodorow measure [5] are some examples of path length based measure. Meaningful similarity measurements based on feature uses more meaningful knowledge than path-length-based method. In addition, feature-based remedies evaluate the difference in the comparison of concepts in generality and ontology, and it is derived from the Tversky similarity model [6] in set theory. The information theory-based method for semantic similarity was first proposed by [7].

Sahni et al. [8] introduced a method in 2014 for measuring semantic similarity of English words. The adopted measures were employed and learned using support vector machines. Jin et al [9] proposed a comprehensive metric of similarity, a method of relatedness measure and a comprehensive degree measure that combines semantic similarity and relatedness between two concepts.

## III. PROBLEM STATEMENT

Technically the problem can be split into
a. Processing the query.
b. How to determine the query type.

c. Determining the class and subclass of the answer of query from the repository.

d. Correlating the semantic similarity of query and predicted answers.

e. Conflict resolution, in case more than one class is predicted.

f. Extraction of answer from the class or sub classes of the repository.

g. Composing more than one sentence if the answer lies with more than one sentence.

## IV.  PROPOSED APPROACH

We are proposing an approach to predict the search result by processing the query using various NLP techniques and then using a series of classifiers to filter into specific portion of the repository where the answer is available. We have taken seven broad classes of sentences as repository, namely:  1. art & culture, 2. Economics, 3. entertainment, 4. Literature, 5. Politics, 6. Sports, and 7. Tourism.

The corpus which has been used in this work is developed under the Technology Development for Indian Languages (TDIL) project of Ministry of Electronics and IT, Govt. of India (Dash 2007) and shared by the Language Research Unit of ISI Kolkata. This corpus which is a size of 11300 A4 size pages, having 271102 numbers of sentences and 3589220 numbers of words, covers 50 different text categories like Agriculture, Child Literature, Physics, Math, Science etc.

The input queries are passed through a set of annotation procedures, like processing of punctuation symbols, uneven spaces, similarization of font, amendment of foreign words by equivalent words in mother language etc. Punctuation marks were taken into account to predict the type of query like Declarative, Imperative, Interrogative or Exclamatory. Then the query is processed to get the parts of speech (POS) of each word using POS tagger of the LTRC, IIIT Hyderabad. The algorithm "Das and Halder" has been used to predict the root form of the verb in the query, tense (present, past or future), person (like 1st, 2nd or 3rd person) etc.

In parallel, our classifier system is trained with already categorized sentences, so that it can predict the category or type of input query correctly. For accurate result four different types of algorithm have been used namely Naive Bayes Probabilistic Model, Support Vector Machine (SMO), Artificial Neural Network (ANN) (multilayer perceptron) and Decision Tree (J48). This process was followed upto *n* level recursively that is upto reaching the level of sub-class to a sentence level atomicity. This model can generate two type of ambiguity.

First, when a same sentence or query is predicted as different categories by different algorithms in a single run, i.e. "Smriti Irani who is a Bollywood actress, came into politics in 2003".  This sentence can be classified as both "art & culture" and "politics". It was resolved based on weighted average i.e. every result of  different algorithms was given a weight of ¼ or 0.25 when a specific category is getting more than 0.50 weight that category is being chosen; if four different category is chosen by four different algorithms (rare case occurred only once) we kept all the predictions.

Second kind of ambiguity is- suppose a sentence is classified as class A by only one algorithm and remaining three algorithms are giving 'NULL prediction' or 'can't be predicted' or only two algorithms are predicting but as two different classes- class A and class B. That means, when the total weight of prediction does not cross 0.50, then we took the only prediction for the first scenario and both of the predictions for the second scenario and passed it for the next level.

Briefly, the algorithms, used for classification are:

### A.  NAIVE BAYES

Subject(s) of a sentence, object(s) of a sentence, Term Frequency, Inverse Document Frequency, length of the text object, dimensionality, entropy, keywords, tense, gender, person, number of subjects and objects, number of Functional Words and Number of Content Words are used as features or attributes for classification using Naive Bayes in our experiment.

As an example, if a Naive Bayes classifier is expected to classify the apples and oranges, then it will apply different attributes of the training set one by one to classify or distinguish them. Suppose it would apply shape first, but apples and oranges both are almost round. Then it would apply color and as apple is red and orange is having different color, it would be able to classify them separately. In case of same color surface texture may be used as attribute as well.

Using Bayes theorem of the conditional probability Naive Bayes classifier classifies objects. Here each object is converted to a vector using word2Vec algorithm. At first the Naive Bayes classifier assumes that all the objects are independent. Then the training is done using the training set and thus the classifier learns to classify the objects. It gives certain numeric value to each object. Those values are the probabilities of the objects to be the member of a certain class based on Bayesian conditional probability. Ultimately the object is assigned to that class in which it gets the maximum probabilistic value.

When a new data point is added, the probabilities are recalculated and adjusted. Assuming that each attribute or feature $x_i$ is independent of any other attribute or feature $x_j$ for j not equals to i, given the category $C_k$ $p(x_i/x_{i+1},…,x_n,C_k) = p(x_i/C_k)$. Thus joint probability written as:

$$p(C_k|x_1, ..., x_n) = p(C_k) \sum_{i=1}^{n} p(x_i|C_k)$$

## B.  SUPPORT VECTOR MACHINE

Support Vector Machine or SVM can classify with both linear and non linear classifier. SVM tries to find out hyper plane with maximum margin between sets of objects. Hyper plane is drawn from the knowledge of training set and its associated vector. The training tuples which fall on the hyper plane is known as support vector. It may also possible that the data are linearly inseparable i.e. with the plane, it's impossible to classify the data objects. In such scenario original input data is transformed into some higher dimensional space using mapping technique and the transformed data in the higher dimensional space becomes separable with hyper plane.

In our experiment first the system was trained with the tagged or classified texts and the SVM classifier model is generated. Then the model is used to classify the test objects with k-1 sets randomly as training set and remaining $k_{th}$ set as testing set. Then the result of classification is tested in average. Subject(s) of a sentence, object(s) of a sentence, Term Frequency, Inverse Document Frequency, length of the text object, dimensionality, entropy, keywords, tense, gender, person, number of subjects and objects, Function Words and Content Words are used as features or attributes for classification using SVM model.

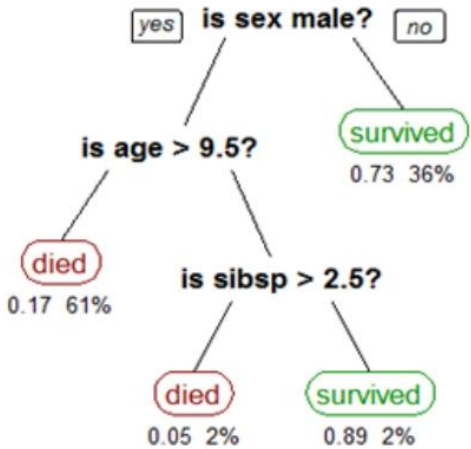

**FIGURE 1.**  Decision Tree

## C.  DECISION TREE

Decision Tree forms a logical representation to take decision in the tree structure under different conditions from the training set. In the training set, first different attributes are identified. Thereafter depending on 'True' and 'False' value of those attributes directly and in nested branches, objects of the training set are categorized in different classes. This predictive model is then used to classify the test data. In the Figure-1, a decision tree is shown which is formed from the data set of passengers after 'The Titanic Mishap'. The classifier has formed a tree with boolean value associated with each branch and nodes or attributes are selected as sex (male - yes or no), age range etc. with the probability of survival. This tree is used to predict the object of a test dataset to determine whether it should survive or not.

In our experiment training data set (9 folds) generates a model or tree with boolean value associated with each branch depending upon various attributes like subject(s) of a sentence, object(s) of a sentence, Term Frequency, Inverse Document Frequency, length of the text object, dimensionality, entropy, keywords, tense, gender, person, number of subjects and objects etc. Then the test (10th) is tested and test is shuffled in various iterations. Decision Tree learns from the percentage value of the training set and applies it as probabilistic value to determine the class of test set.

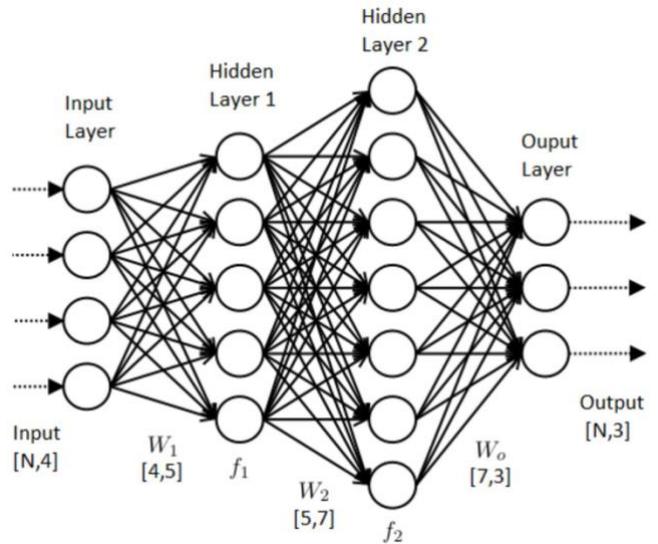

**FIGURE 2.**  Artificial Neural Network

## D.  ARTIFICIAL NEURAL NETWORK

Conceptualized to mimic, the learning pattern of brain and to improve the learning rate over numbers of iterations, Artificial Neural Network is designed. From 2012 it started to get huge popularity with the multilayered feed forward network and recurrent neural network and advanced, scalable, distributed GPUs and thus initializing deep learning by automating the load of feature extraction. In different machine learning challenges like speech recognition, pattern recognition it showed almost 15 to 20 percent improvement with respect to traditional statistical methods.

ANN takes considerably long time to learn from the input data set depending upon the size of the dataset. The main advantage of ANN is- it can correct itself to improve the accuracy with the number of iteration. There are input layers, hidden layers and output layers. Where input layers get direct input and output of the output layer is treated as the final output of the ANN. Hidden layers don't get direct input from the outside environment rather it takes input from other layers of the ANN typically other hidden layer or input layer. Number of hidden layers also depends upon the design.

**Algorithm 1: DAS & HALDER**

**Input:** Bengali corpus to Shallow parser (LTRC)

**Output:** Inflected verbs are collected in a file Input.doc

**Input to our system:** Input.doc

**Output of our system:** Multiple files classified according to tense and person consisting of root form of verbs

1  begin
2  | Corpus is taken as input to the Shallow parser
3  | All the inflected verbs are collected in a file named as Input.doc
4  | Input.doc is taken as input to our system
5  | if বিভক্তি(in the Table 1) matches then
6  |   | if *person matches* then
7  |   |   | Verbs are passed to the respective method with the verbs and বিভক্তি for processing
8  |   |   | Verbs are processed one by one
9  |   |   | Panini's rules are applied on those verbs to extract root verbs
10 |   |   | Root verb is stored in Output.doc file as well as separate file according to tense.
11 |   | else
12 |   |   | The verb collected from Shallow parser in Input.doc is not a verb
13 | else
14 |   | The verb collected from Shallow parser in Input.doc is not a verb

**FIGURE 3.  Das and Halder Algorithm**

## V. METHODOLOGY

1. Call the Shallow Parser to get the Parts of Speech (POS) of each words of the input query.

2. Mark Function Words and Content Words in the input sentence.

3. Use "Das and Halder" algorithm to extract the root form of the verb [Fig. 3]

4. Use WordNet to get the synonyms and to enhance the dimensionality of the input vector.

5. Classify the query to one of the seven classes using Naive Bayes, ANN, SVM and Decision Tree algorithm.

6. Apply Step-5 recursively to determine the subclass.

7. Hit the target sentences and use knowledge base to extract the answer in desired format.

8. Return the result to the user.

The detailed flowchart of our work is depicted in Figure-5. Use of Shallow Parser and Extraction of Root Verb are two separate works, detail of which has not been described in the flowchart. Shallow Parser developed by LTRC group of IIIT Hyderabad, is used and thus the algorithm of POS tagging is not covered in this paper, acknowledgement has been given at last of this paper and IIIT Hyderabad has also been informed about the percentage of improvement in the result because of the POS tagger. Features like Function Words(FW), Content Words(CW), number of FW and number of CW, subjects, objects and their numbers are extracted using Shallow Parser.

The detail of Root Verb Extraction algorithm is given in the Das and Halder Algorithm (Figure 3) which is actually used to extract the feature like person, number, gender, tense of subjects and objects.

We have used the Java language for implementation of supervised algorithm for automatic root verb extraction.

Weka was used for different classification algorithm usage.

Java API has been used to call it recursively.

PostgreSQL relational database has been used as knowledge base.

Training set was prepared by the researchers without getting the test set. Model was generated then using that model and training set test set was evaluated. Here test set was generated from the user query.

At each stage the result was evaluated by the method of cross-validation. Then the predicted sentence was passed to the "extractor" program which with the help of knowledge base formed the answer. The same was returned to the user and his satisfaction was recorded to measure the performance of our system.

WordNet has been used in the two stages, first to make our system understand the user query whenever required. Mostly when the meaning of any particular word is not known, our system tries to replace those words with the entries from word net.

Secondly during the formation of answer to return the result set to the user, WordNet is being used again to form the accurate answer and also in case of ambiguity to give more than one context of answers.

Some of the indexing settings we have used in weka tools, used for classification are -

| Tense | Type of tense | Suffices (বিভক্তি) |
|-------|---------------|---------------------|
| বর্তমান কাল (Present Tense) | সামান্য বর্তমান Simple Present | "ি","ে","েন","িস", "ই" |
| | ঘটমান বর্তমান Present Continuous | "ছ্","িতেছ্","ছে" ,"িতেছে" ,"ছ্","িতেছ","ছেন" |
| | পুরাঘটিত বর্তমান Present Perfect | "েছি","িয়াছি","েছ","িয়াছ", "েছে","েছেন" |
| অতীত কাল (Past Tense) | সামান্য অতীত Simple Past | "লাম","লুম","িলাম","িলুম","লে","িলে","লেন","িলেন" |
| | ঘটমান অতীত Past Continuous | "ছিলাম","ছিলুম","িতেছিলাম","িতেছিলুম" |
| | পুরাঘটিত অতীত Past Perfect | "েছিলাম","েছিলুম","িয়াছিলাম" |
| | নিত্যবৃত্ত অতীত Past Perfect Continuous | "তাম","তুম","িতাম","িতুম","তে" |
| ভবিষ্যৎ কাল (Future Tense) | সামান্য ভবিষ্যৎ Simple Future | "ব","িব","বে","িবে","বি","িবি" |
| | ঘটমান ভবিষ্যৎ Future Continuous | "তেথাকব","িতেথাকিব,তেথাকিবে","িতেথাকিবি","তেথাকিবি", |
| | পুরাঘটিত ভবিষ্যৎ Future Perfect | "েথাকব","িয়াথাকিব", "েথাকিবে" |

**FIGURE 4. Table1 : Different kind of suffices applied to the verb in Bengali with Tenses**

**A. USED TRAINING SET**
It is used to train the system as well as it is used to evaluate the model of the classifier how well it classifies the training set itself.

**B. SUPPLIED TEST SET**
It is the test set which we actually want to classify. It is used to evaluate the predictive performance of the classifier.

**C. CROSS VALIDATION**
Cross validation is a technique to make an average of the test result. A dataset is split into X sets or pieces ("folds"). Then X-1 sets are used for training and remaining Xth set is used for testing. It gives X evaluation results and they are averaged. In case of 2 fold validation, the dataset is divided into d0 and d1, both of which are of equal size. We first train the system with d0 and test d1 then train the system with d1 and test d0. The result is then summed up and divided by 2.We have taken here 10 folds cross validation.

**D. PERCENTAGE SPLIT**
It mentions the portion of the data which is used as training and remaining is used as test data. Suppose percentage split is 70 percent and there are total 100 instances of the data. Then from 0th to 69th instances of the dataset is used as training and 70th to 99th dataset is used as test data. Random split is used with the help of seed value.

**E. OUTPUT MODEL**
Output model is generated based on training set. It can be visualized and verified.

**F. OUTPUT PER CLASS STATS**
For every class output this is the precision/recall and true/false statistics.

**G. OUTPUT CONFUSION MATRIX**
The confusion matrix is the one of the key metric to test the performance of the classifier.

**H. SCORE PREDICTION FOR VISUALIZATION**
The classifier's predictions are remembered so that they can be visualized.

**I. RANDOM SEED FOR X VAL / PERCENTAGE SPLIT**
This specifies the random seed used when randomizing the data before it is divided up for evaluation purposes.

**J. OUTPUT ENTROPY EVALUATION MEASURE**
Entropy evaluation measures are included in the output.

**K. OUTPUT PREDICTION**
The classifier's predictions are remembered, so that they can be visualized.

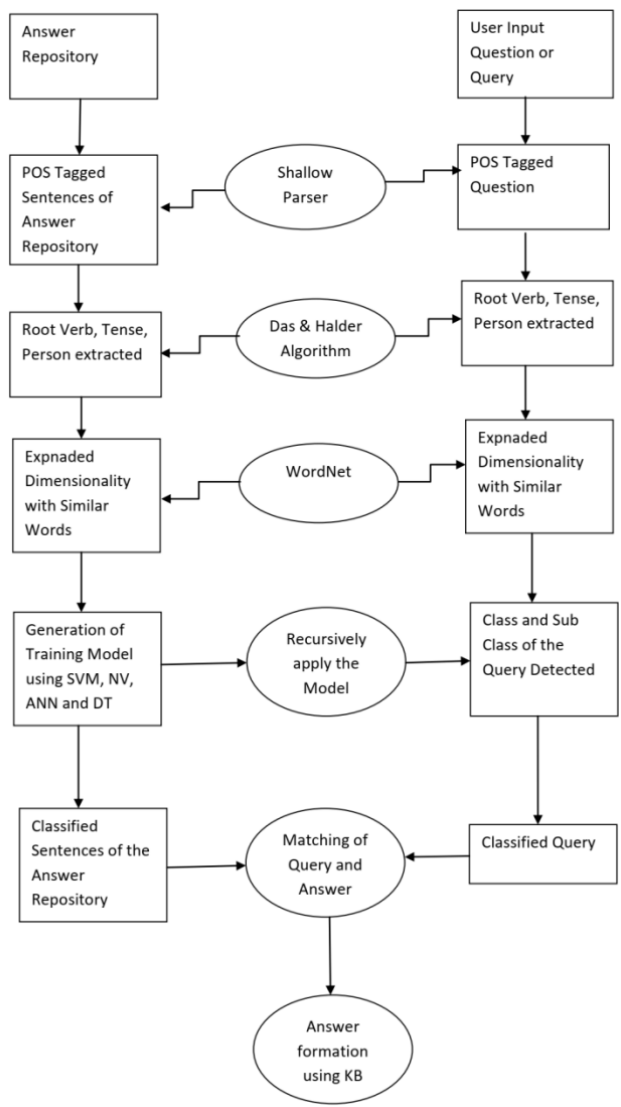

**FIGURE 5.** Detailed Flowchart of the Methodology Used

## VI.  RESULT

### A.  RESULT SUMMARY

98 percent accuracy is achieved in the Root Verb Extraction by Das & Halder Algorithm [10]. Confusion matrix for four different classification methods is given in Table 2, Table 3, Table 4 and Table 5.

A confusion matrix is a summary of prediction results on a classification problem. The number of correct and incorrect predictions are summarized with count values and broken down by each class. This is the key to the confusion matrix. The confusion matrix shows the ways in which classification model is confused when it makes predictions. It gives us insight not only into the errors being made by a classifier but more importantly the types of errors that are being made.

Out of 250 questions for 244 questions the system has hit the correct sentence(s) where the answer is hidden giving 97.6 percent for hit success.

Out of 250 questions for 214 questions the answers were given accurate by the system with 85.6 percent answer accuracy with moderate grammatical correctness.

### B.  DETAILED RESULT

#### i) Naïve Bayes
Correctly Classified Instances -88 percent
Incorrectly Classified Instances-12 percent
Kappa statistic-0.1961
Mean absolute error-0.4664
Root mean squared error-0.5008
Relative absolute error-93.3564 percent
Root relative squared error-100.1666 percent
Total Number of Instances-100

So our model identifies 88 correctly Classified Instances and, 12 incorrectly classified instances. So the accuracy of the model is 88 percent and the inaccuracy of the model is 12 percent.
Confusion matrix for Naive Bayes is given in table 1.

#### ii) SVM (SMO)
Correctly Classified Instances-86 percent
Incorrectly Classified Instances-14 percent
Kappa statistic-0.2834
Mean absolute error-0.36
Root mean squared error-0.6
Relative absolute error-72.0627 percent
Root relative squared error-120.0079 percent
Total Number of Instances-100

So our model identifies 86 correctly classified instances and 14 incorrectly classified Instances. The accuracy of the model is 86 percent   and the inaccuracy of the model is 14 percent.
Confusion matrix for SVM is given in table 2.

#### iii) ANN
Correctly Classified Instances-96 percent
Incorrectly Classified Instances-4 percent
Kappa statistic-0.1186
Mean absolute error-0.442
Root mean squared error-0.59
Relative absolute error-8.478 percent
Root relative squared error-118.0054 percent
Total Number of Instances-100

So our model identifies 96 Correctly Classified Instances and 4 incorrectly classified Instances. The accuracy of the model is 96 percent and inaccuracy of the model is 4 percent.
Confusion matrix for SVM is given in table 3.

#### iv) DECISION TREE
Correctly Classified Instances-73 percent
Incorrectly Classified Instances-27 percent
Kappa statistic-0.4534

Mean absolute error-0.3493
Root mean squared error-0.4748
Relative absolute error-69.9227 percent
Root relative squared error-94.9762 percent
Total Number of Instances-100

So our model identifies 73 Correctly Classified Instances and 27 incorrectly classified Instances. The accuracy of the model is 73 percent and inaccuracy of the model is 27 percent.

Confusion matrix for Decision Tree is given in table 4.

## VII.  PERFORMANCE ANALYSIS

Detailed analysis of result is

### A.  Naive Bayes

===Runinformation===
Scheme:weka.classifiers.bayes.NaiveBayes
Relation:weka.datagenerators.classifiers.classification.
Instances:100
Attributes:10
class
Node2
Node3
Node4
Node5
Node6
Node7
Node8
Node9
Node10

Test mode: split 66.0 percent train, remainder test

=== Classifier model (full training set) ===
Naïve_Bayes_Classifier
Class

| Attribute | Value1 (0.52) | Value2 (0.48) |
|---|---|---|
| class | | |
| Value1 Value2 [total] Node2 | 23.0 31.0 54.0 | 29.0 21.0 50.0 |
| Value1 Value2 [total] Node3 | 33.0 21.0 54.0 | 27.0 23.0 50.0 |
| Value1 Value2 [total] Node4 | 22.0 32.0 54.0 | 26.0 24.0 50.0 |
| Value1 Value2 [total] | 26.0 28.0 54.0 | 22.0 28.0 50.0 |

| Node5 | | |
|---|---|---|
| Value1 Value2 [total] Node6 | 27.0 27.0 54.0 | 22.0 28.0 50.0 |
| Value1 Value2 [total] Node7 | 20.0 34.0 54.0 | 24.0 26.0 50.0 |
| Value1 Value2 [total] Node8 | 27.0 27.0 54.0 | 39.0 11.0 50.0 |
| Value1 Value2 [total] Node9 | 22.0 32.0 54.0 | 25.0 25.0 50.0 |
| Value1 Value2 [total] | 29.0 25.0 54.0 | 29.0 21.0 50.0 |

Time taken to build model: 0.05 seconds
=== Evaluation on test split ===
Time taken to test model on test split: 0.05 second

### B.  SVM(SMO)
=== Run information ===
Scheme:weka.classifiers.functions.SMO
Relation:      weka.datagenerators.classifiers.classification.
Instances:                                               100
Attributes:                                               10
class
Node2
Node3
Node4
Node5
Node6
Node7
Node8
Node9
Node10
Test mode: split 66.0 percent train, remainder test
=== Classifier model (full training set) ===
SMO
Kernel                                              used:
Linear        Kernel:        K(x,y)        =        <x,y>
Classifier    for    classes:    Value1,    Value2
Binary                                              SMO
Machine linear: showing attribute weights, not support vectors.
-0.5999        *        (normalized)        class=Value2
+      -0.0002        *        (normalized)        Node2=Value2
+      -0.6005        *        (normalized)        Node3=Value2
+      0.0004        *        (normalized)        Node4=Value2
+      0.5997        *        (normalized)        Node5=Value2
+      0.0003        *        (normalized)        Node6=Value2

| | | | |
|---|---|---|---|
| + | -1.3999 | * | (normalized) Node7=Value2 |
| + | -0.5998 | * | (normalized) Node8=Value2 |
| + | 0.0005 | * | (normalized) Node9=Value2 |
| + | | | 0.9999 |

Number of kernel evaluations: 3284 (80.111 percent cached)
Time taken to build model: 0.05 seconds
=== Evaluation on test split ===
Time taken to test model on test split: 0.06 second

### C.  Multi Layer Perceptron (ANN)

=== Run information ===
Scheme:weka.classifiers.functions.MultilayerPerceptron
Relation:     weka.datagenerators.classifiers.classification.
Instances:                                          100
Attributes:                                          10
class
Node2
Node3
Node4
Node5
Node6
Node7
Node8
Node9
Node10

Test mode: split 86.0 percent train, remainder test
=== Classifier model (full training set) ===

| Sigmoid | Node | | 0 |
|---|---|---|---|
| Inputs | | | Weights |
| Threshold | | | -0.46337480146533117 |
| Node | 2 | | 8.103745556620964 |
| Node | 3 | | -6.072875619653186 |
| Node | 4 | | 6.372495140324609 |
| Node | 5 | | -3.9503308596408644 |
| Node | 6 | | -7.215969568907012 |

| Sigmoid | Node | | 1 |
|---|---|---|---|
| Inputs | | | Weights |
| Threshold | | | 0.4633740069800474 |
| Node | 2 | | -8.1034974021937 |
| Node | 3 | | 6.072668908499916 |
| Node | 4 | | -6.372283394108063 |
| Node | 5 | | 3.950231863927222 |
| Node | 6 | | 7.215717271685786 |

| Sigmoid | Node | | 2 |
|---|---|---|---|
| Inputs | | | Weights |
| Threshold | | | 2.7683245856190135 |
| Attrib | class=Value2 | | 0.8682762442270585 |
| Attrib | Node2=Value2 | | -4.431137559546951 |
| Attrib | Node3=Value2 | | 3.783617770354907 |
| Attrib | Node4=Value2 | | -4.380158025756841 |
| Attrib | Node5=Value2 | | -6.569100948175245 |
| Attrib | Node6=Value2 | | 1.2644761203361334 |
| Attrib | Node7=Value2 | | 10.136894593720866 |
| Attrib | Node8=Value2 | | 2.2222931644808117 |
| Attrib | Node9=Value2 | | -1.1961756750583683 |

| Sigmoid | Node | | 3 |
|---|---|---|---|
| Inputs | | | Weights |
| Threshold | | | 1.3449330093589562 |
| Attrib | class=Value2 | | -1.4873176018858711 |
| Attrib | Node2=Value2 | | -3.4738558540312514 |
| Attrib | Node3=Value2 | | -0.8903481429148983 |
| Attrib | Node4=Value2 | | -3.30524117606012 |
| Attrib | Node5=Value2 | | -1.9954254971320733 |
| Attrib | Node6=Value2 | | -5.043053347446121 |
| Attrib | Node7=Value2 | | -7.272891242499204 |
| Attrib | Node8=Value2 | | 3.48425700470805 |
| Attrib | Node9=Value2 | | -5.422571254947003 |

| Sigmoid | Node | | 4 |
|---|---|---|---|
| Inputs | | | Weights |
| Threshold | | | 0.869193799614438 |
| Attrib | class=Value2 | | 5.390911683768276 |
| Attrib | Node2=Value2 | | -0.7486372607912477 |
| Attrib | Node3=Value2 | | -6.390427323436479 |
| Attrib | Node4=Value2 | | -3.6222767041324264 |
| Attrib | Node5=Value2 | | -4.135964801658738 |
| Attrib | Node6=Value2 | | -5.691365511528745 |
| Attrib | Node7=Value2 | | 2.727486953876179 |
| Attrib | Node8=Value2 | | 0.6095482877748186 |
| Attrib | Node9=Value2 | | -1.7878361719212885 |

| Sigmoid | Node | | 5 |
|---|---|---|---|
| Inputs | | | Weights |
| Threshold | | | -3.764934366886755 |
| Attrib | class=Value2 | | 0.8927457455455018 |
| Attrib | Node2=Value2 | | -1.9686974609437013 |
| Attrib | Node3=Value2 | | -2.295112500524242 |
| Attrib | Node4=Value2 | | -4.21750653575111 |
| Attrib | Node5=Value2 | | -3.2436685768106672 |
| Attrib | Node6=Value2 | | -2.333295775325579 |
| Attrib | Node7=Value2 | | 3.217524336891349 |
| Attrib | Node8=Value2 | | -0.9017912544548131 |
| Attrib | Node9=Value2 | | 0.7290063727690463 |

| Sigmoid | Node | | 6 |
|---|---|---|---|
| Inputs | | | Weights |
| Threshold | | | -4.36151269195419 |
| Attrib | class=Value2 | | -2.535835981153719 |
| Attrib | Node2=Value2 | | 0.5990975373610684 |
| Attrib | Node3=Value2 | | -1.9828745134956305 |
| Attrib | Node4=Value2 | | -3.7293430187368592 |
| Attrib | Node5=Value2 | | -3.208820300338705 |
| Attrib | Node6=Value2 | | 0.6159441682751483 |
| Attrib | Node7=Value2 | | 7.332032494106006 |
| Attrib | Node8=Value2 | | -0.1418907230980749 |
| Attrib | Node9=Value2 | | -0.23822839374741406 |

| Class | | | Value1 |
|---|---|---|---|
| Input | | | |
| Node | | | 0 |
| Class | | | Value2 |
| Input | | | |
| Node | | | 1 |

Time taken to build model: 0.1 seconds
=== Evaluation on test split ===
Time taken to test model on test split: 0.05 seconds

### D.  Decision Tree J48

```
===           Run        information         ===
Scheme:weka.classifiers.trees.J48
Relation:      weka.datagenerators.classifiers.classification.
Instances:                                             100
Attributes:                                             10
class
Node2
Node3
Node4
Node5
Node6
Node7
Node8
Node9
Node10
Test mode: split 66.0 percent train, remainder test
===   Classifier   model   (full   training   set)   ===
J48                  pruned                  tree
------------------
Node7                     =                  Value1
|          Node5              =              Value1
|    |    Node8   =   Value1:   Value1   (15.0/3.0)
|    |    |        Node8       =       Value2
|    |    |    Node2  =  Value1:  Value1  (7.0/2.0)
|    |    |    Node2  =  Value2:  Value2  (12.0/2.0)
|         Node5           =           Value2
|    |    Node8  =  Value1:   Value2   (15.0)
|    |    |        Node8       =       Value2
|    |    |    |       Node4      =      Value1
|    |    |    |    class  =  Value1:  Value2  (3.0/1.0)
|    |    |    |    class  =  Value2:  Value1  (6.0/1.0)
|    |    |    Node4  =  Value2:  Value1  (6.0/1.0)
Node7                     =                  Value2
|          Node4              =              Value1
|    |    |       Node5       =       Value1
|    |    |    Node9  =  Value1:  Value2  (4.0)
|    |    Node9  =  Value2:  Value1  (4.0/1.0)
|    |    |       Node5       =       Value2
|    |    |    |       class      =      Value1
|    |    |    Node2  =  Value1:  Value2  (2.0)
|    |    |    Node2  =  Value2:  Value1  (3.0/1.0)
|    |    |    class  =  Value2:  Value1  (11.0/1.0)
|     Node4  =  Value2:  Value1  (12.0/1.0)
Number       of        Leaves      :         13
Size      of       the      tree      :         25
Time   taken   to   build   model:   0.02   seconds
===    Evaluation    on    test    split    ===
Time taken to test model on test split: 0.01 seconds.
```

## VIII.  APPLICATIONS

Semantic Search improves the contextual meaning finding. When user is asking some question to get the answer, it plays a crucial role. Thus for automatic question answering system, semantic search is the backbone.

For geographical map annotation semantic search is extensively used.

The knowledge acquired from semantic search in text analytics are being used in bio informatics as well.

It has a huge importance in Automatic Question Answering system, News Classification, Text Summarization, WordNet improvement, Sentiments Analysis.

## IX.  SCOPE FOR IMPROVEMENTS

Making the Knowledge Base a self learning system is the next challenge. This is possible by the way of semi-supervised learning incorporating the human intelligence into the system. Where the output of the system will be verified by the human feedback and in the next iteration it will improve by cognition.

The algorithms used, are completely language independent. The performance of the system has been tested in Indo Aryan Language groups. To test the same for other language groups are the next possible enhancements.

## X.  CONCLUSION

In this work, an attempt is made to design an effective algorithm for semantic search. There are major three tasks, first is to process the query and the second is to point the portion of the repository where the probable answer is hidden and the third method is to frame the answer from the pointed sentences. POS tagging, Root verb extraction, recursive classification to predict the portion of repository where probable answer is hidden and at the last stage, extracting the answer using knowledge base has been used. Artificial Neural Network, Naive Bayes, SVM (SMO) and Decision Tree have been used as statistical process to classify. Finally the knowledge base was used to form the answer and return to the user.

At the initial stage, the testing of the performance of the system was done on the Indic Language dataset like Bengali. The accuracy was measured by the expert linguists. From the beginning the design and development of the system was carried out in such a fashion that it can be useful globally without any regional language constraint. The same has been tested at later stage with other languages also and the system is running perfectly well. Measuring the accuracy, precision for all the natural languages in the world is beyond our capability and it is expected that the researchers and the linguists of other language groups will use our algorithm and compare the performance with their own system.

As of now no question answering system is available for Bengali. Most of them are available in English or popular European languages. So our research work is aimed particularly to cater the necessity of the mankind of that language group which has low resource, low popularity but this work also in general applicable to universally any language.

## ACKNOWLEDGMENT

We are grateful to LTRC group, IIIT Hyderabad for providing Shallow Parser for POS tagging.
We are also thankful to Professor Niladri Sekhar Das of Indian Statistical Institute, Kolkata (ISI Kolkata) for providing the dataset. He being the renowned linguist also helped in evaluation.

## AUTHORS PROFILE

**ARIJIT DAS** received B.Tech. degree in Computer Science and Engineering in 2011 from Govt. College of Engineering and M.E. in Computer Science and Engineering in 2013 from Jadavpur University, Kolkata, India with GATE fellowship. Then he joined as Scientific Officer in the Ministry of IT, Govt. of India. Currently he is pursuing PhD (Engg.) in Jadavpur University. He became the member of IEEE in 2016.

**DIGANTA SAHA** is currently working as Professor in the Department of Computer Science and Engineering in Jadavpur University. He works in the field of Natural Language Processing.